# Medical artificial intelligence toolbox (MAIT): an explainable machine learning framework for binary classification, survival modelling, and regression analyses

Ramtin Zargari Marandi (1), Anne Svane Frahm (1), Jens Lundgren (1), Daniel Dawson Murray (1), Maja Milojevic (1)

(1) Centre of Excellence for Health, Immunity and Infections (CHIP), Rigshospitalet, Copenhagen University Hospital, Blegdamsvej 9, DK-2100 Copenhagen, Denmark.

While machine learning offers diverse techniques suitable for exploring various medical research questions, a cohesive synergistic framework can facilitate the integration and understanding of new approaches within unified model development and interpretation. We therefore introduce the Medical Artificial Intelligence Toolbox (MAIT), an explainable, open-source Python pipeline for developing and evaluating binary classification, regression, and survival models on tabular datasets. MAIT addresses key challenges (e.g., high dimensionality, class imbalance, mixed variable types, and missingness) while promoting transparency in reporting (TRIPOD+AI compliant). Offering automated configurations for beginners and customizable source code for experts, MAIT streamlines two primary use cases: Discovery (feature importance via unified scoring, e.g., SHapley Additive exPlanations - SHAP) and Prediction (model development and deployment with optimized solutions). Moreover, MAIT proposes new techniques including fine-tuning of probability threshold in binary classification, translation of cumulative hazard curves to binary classification, enhanced visualizations for model interpretation for mixed data types, and handling censoring through semi-supervised learning, to adapt to a wide set of data constraints and study designs. We provide detailed tutorials on GitHub, using four open-access data sets, to demonstrate how MAIT can be used to improve implementation and interpretation of ML models in medical research.

## Introduction

Machine learning (ML) has demonstrated an ability to enhance data processing across various domains including medicine[1,2]. Despite the advancement of ML methods and algorithms, there are still important factors creating gaps in the utilization of ML for clinical research. A hindering factor in clinical research may be attributed to the complexities in implementation of ML methodologies tailored to working with medical datasets and study designs[3]. In contrast, conventional statistical approaches, like logistic regression and Cox models are popular within the medical community due to the extensive historical usage of terminologies and methodologies ingrained over many years.

The rise in popularity of ML can be attributed to several factors. One reason is the ability of ML to incorporate both well-established statistical models like logistic regression, and novel algorithms (e.g., LightGBM[4]), to extract impactful information and solve complex problems. In addition, ML models are often known for their data-driven approach and are less reliant on strong assumptions as compared to some statistical models. Another useful aspect of ML models is the ability to continuously learn from new data to mitigate the risk of model drift over time, which is important in cases such as the use of digital tools for patient screening. The rise in more wide-spread ML use also coincides with the increasing availability of computational infrastructure (e.g., high-performance computing clusters and graphics processing units - GPUs) which is needed to process the ever-growing amounts of big data[5].

In medical research, non-image patient data is often organized in tabular formats with features or risk factors as columns and individuals as rows. Binary classification within these data structures serves two critical roles: 1) discovery - where patterns and predictive relationships between variables are identified, and 2) prediction – which focuses on building deployment-ready models to forecast outcomes from current health indicators. Prediction models can be used as diagnostic tools to aid in immediate clinical interventions, or as prognostic indicators, anticipating future medical needs. For time-to-event data, survival analysis using methods like random survival forests offers an alternative analytical approach that accounts for censoring and avoids the potential biases inherent to standard binary classification approaches.

MAIT is a highly versatile tool allowing users to address the multifaceted challenges of ML applications within medical research, particularly when confronted with high-dimensional datasets with mixed variable types and missing values. It provides an accessible and customizable ML framework for developing models with varying levels of user expertise while maintaining transparency aligned with the TRIPOD+AI statement[6]. MAIT automates many data preprocessing and modeling steps while allowing for customization of the processes within a standard framework. In addition, MAIT bridges the analytical gap between survival analyses and binary classification, an integral aspect for clinical decision-making based on ML analyses. The novelty of MAIT relies on its implemented techniques used to facilitate the preprocessing of tabular data for compatibility for a wide variety of algorithms, to provide comprehensive model evaluation and validation, and allow for unified and multi-level model interpretation. Consequently, this consolidation of methodologies into one cohesive synergistic system boosts the potential to clarify complex ML applications in medical research. Our aims are therefore to provide a semi-automated framework in a form of an explainable ML pipeline, where there is a reasonable level of human involvement in ML design[7] while maintaining a consistent framework for model evaluation and reporting, as well as addressing some common technical constraints.

## Overview

The current version of MAIT, v. 1.0.0, is implemented in a Jupyter Notebook using Python packages (modules) commonly employed in ML tasks like Scikit-Learn[8]. MAIT is executable within a Conda environment from Anaconda Software Distribution (Anaconda Inc.) as instructed in MAIT documentations (publicly available on GitHub, https://github.com/PERSIMUNE/MAIT, containing instructions, tutorials for common use cases, and details on all the methods and algorithms). The environment to execute MAIT is also available as a Docker Image (Docker Inc.). Here, we provide a summary of the features and functionalities of MAIT (Figure 1), as well as additional documentation thoroughly describing the technical details.

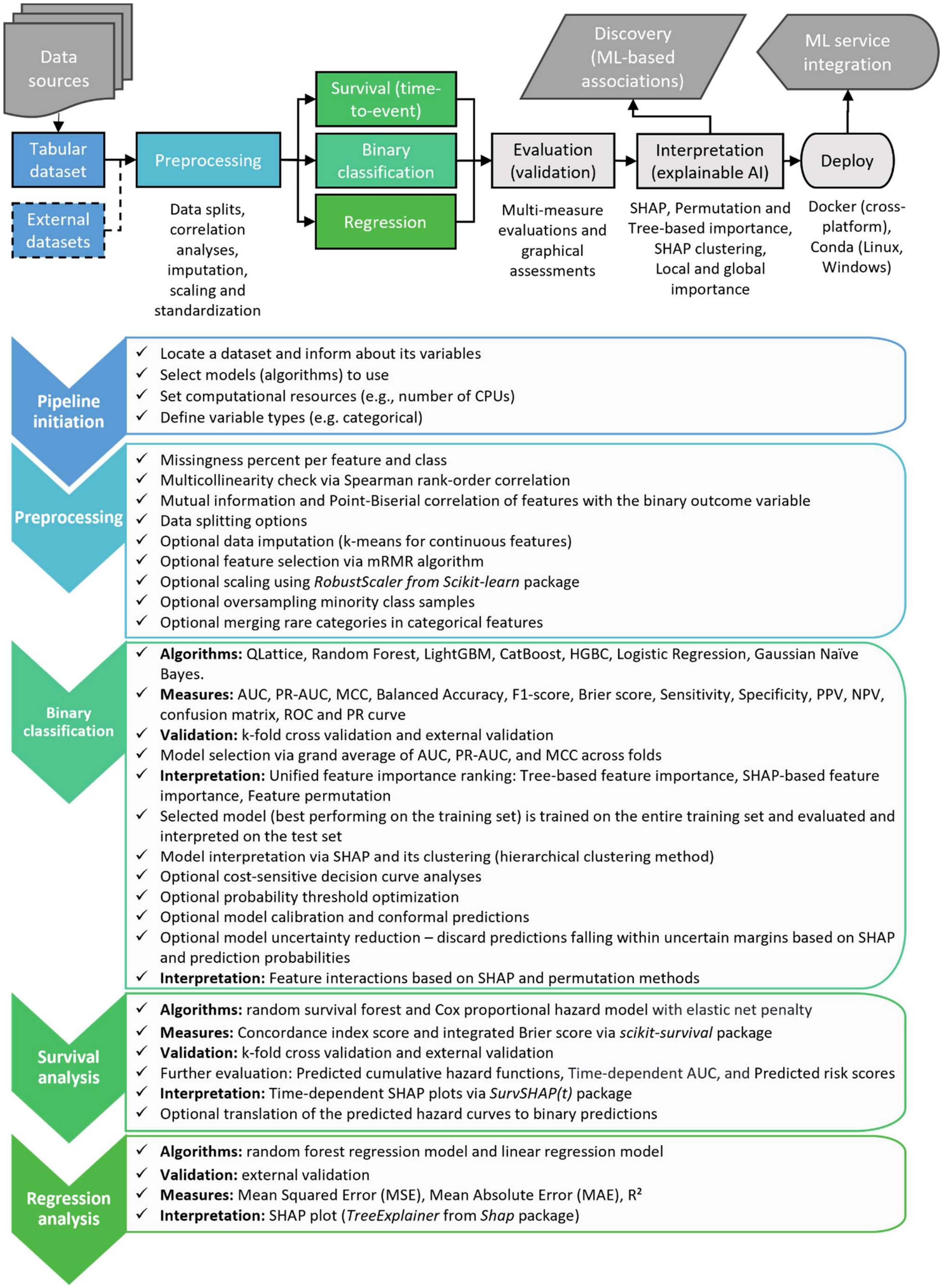
Data sources
Tabular dataset
External datasets
Preprocessing
Data splits, correlation analyses, imputation, scaling and standardization
Survival (time-to-event)
Binary classification
Regression
Evaluation (validation)
Multi-measure evaluations and graphical assessments
Interpretation (explainable AI)
SHAP, Permutation and Tree-based importance, SHAP clustering, Local and global importance
Discovery (ML-based associations)
Deploy
Docker (cross-platform), Conda (Linux, Windows)
ML service integration
Pipeline initiation
✓ Locate a dataset and inform about its variables
✓ Select models (algorithms) to use
✓ Set computational resources (e.g., number of CPUs)
✓ Define variable types (e.g. categorical)
Preprocessing
✓ Missingness percent per feature and class
✓ Multicollinearity check via Spearman rank-order correlation
✓ Mutual information and Point-Biserial correlation of features with the binary outcome variable
✓ Data splitting options
✓ Optional data imputation (k-means for continuous features)
✓ Optional feature selection via mRMR algorithm
✓ Optional scaling using RobustScaler from Scikit-learn package
✓ Optional oversampling minority class samples
✓ Optional merging rare categories in categorical features
Binary classification
✓ Algorithms: QLattice, Random Forest, LightGBM, CatBoost, HGBC, Logistic Regression, Gaussian Naïve Bayes.
✓ Measures: AUC, PR-AUC, MCC, Balanced Accuracy, F1-score, Brier score, Sensitivity, Specificity, PPV, NPV, confusion matrix, ROC and PR curve
✓ Validation: k-fold cross validation and external validation
✓ Model selection via grand average of AUC, PR-AUC, and MCC across folds
✓ Interpretation: Unified feature importance ranking: Tree-based feature importance, SHAP-based feature importance, Feature permutation
✓ Selected model (best performing on the training set) is trained on the entire training set and evaluated and interpreted on the test set
✓ Model interpretation via SHAP and its clustering (hierarchical clustering method)
✓ Optional cost-sensitive decision curve analyses
✓ Optional probability threshold optimization
✓ Optional model calibration and conformal predictions
✓ Optional model uncertainty reduction – discard predictions falling within uncertain margins based on SHAP and prediction probabilities
✓ Interpretation: Feature interactions based on SHAP and permutation methods
Survival analysis
✓ Algorithms: random survival forest and Cox proportional hazard model with elastic net penalty
✓ Measures: Concordance index score and integrated Brier score via scikit-survival package
✓ Validation: k-fold cross validation and external validation
✓ Further evaluation: Predicted cumulative hazard functions, Time-dependent AUC, and Predicted risk scores
✓ Interpretation: Time-dependent SHAP plots via SurvSHAP(t) package
✓ Optional translation of the predicted hazard curves to binary predictions
Regression analysis
✓ Algorithms: random forest regression model and linear regression model
✓ Validation: external validation
✓ Measures: Mean Squared Error (MSE), Mean Absolute Error (MAE), R²
✓ Interpretation: SHAP plot (TreeExplainer from Shap package)

Figure 1 Overview of the analytical blocks of MAIT. It includes several methods and optional functions to address common methodological challenges for binary classification, survival, and regression analyses. The pipeline begins with user-defined settings selecting the methods to use and providing information about the input data. First, the data is imported and quality checked, then MAIT's core functionality of binary classification is applied. The pipeline also allows for survival and regression modeling. The blocks are in a sequence of code chunks within a Jupyter Notebook file for readability. For a complete list of all MAIT functionalities see https://github.com/PERSIMUNE/MAIT. Acronyms: area under the receiver operating characteristics curve (AUC), area under the precision recall curve (PR-AUC), Matthews correlation coefficient (MCC), positive predictive value (PPV), negative predictive value (NPV), SHapley Additive exPlanations (SHAP), and histogram-based gradient boosting classifier (HGBC).

## Pipeline initiation

For pipeline initiation, analysis metadata (e.g., data localization, definition of variables, methods and algorithms to use) and computational resources (e.g., GPU usage) must first be configured. There are also extra options available to the user, such as enabling feature selection and its detailed options (e.g., number of features). MAIT requires datasets for ML analyses to be structured as a tabular dataset where each row is an observation (e.g., patient), and each column is a variable, such as a patient characteristics like age or gender. Such formats are common for electronic health records stored as tab- or comma-separated values and excel files. In binary classification, a dataset is required to have a column for the outcome variable in a binary (i.e., two categories). In case of outcome censoring (e.g., lost to follow up), semi-supervised learning, through a label propagation method[9], is available to address relabeling of instances with unknown labels.

For diagnosis or prognosis, validation on an independent dataset, usually called a test set, is necessary. The option for data splitting of a single dataset into development and testing sets is therefore available in MAIT. Alternatively, two separate, independent datasets can be used for development and testing. MAIT ensures that all data processing steps are performed independently (in isolation) for each dataset (i.e., development or train(ing), test(ing), and external validation sets) to avoid data leakage[10]. In case without data splitting, the pipeline can serve for association studies based on ML models for discovery purposes rather than for deployment of models for prediction.

Data representativeness is crucial for fair assessments of model performance and generalizability[11]. Depending on the data, there may be multiple criteria that need consideration to ensure that both the development and test sets adequately represent the underlying population. For instance, if one population sub-group or patient characteristic accounts for 10% of all samples, it is important to maintain a similar proportion in both sets. In such cases, if the outcome, such as mortality with a prevalence of 20%, is considered, the two factors can be combined in MAIT to form a composite feature for stratified data splitting.

The pipeline checks data quality for missingness, rare categories, and feature variance. MAIT also assesses correlation of the features to each other and the binary outcome variable by Spearman's rank correlation. Spearman's rank correlation is often preferred over Pearson's correlation due to its ability to capture non-linear relationships, be more robust to outliers, and not rely on assumptions of normality. To complement Spearman's rank correlation using a non-ranking method, Point-Biserial correlation and mutual information are used to assess the correlation of all features with the binary outcome. MAIT also allows filtering out of highly missing instances and features, merging highly rare categories in categorical features, and outlier detection via the isolation forest algorithm[12]. The data exploration analyses, performed prior to ML, provide preliminary assessments of the data distribution and the strength of associations in the data.

MAIT includes algorithms that can handle missingness and categorical features, like CatBoost[13] and LightGBM[4], as well as algorithms that require some data transformation. As such, one-hot encoding is applied for categorical variables to transform them into binary variables and k-nearest neighbors is used to impute missingness in continuous variables (*KNNImpute* function in *Scikit-Learn*). An optional data scaling method based on interquartile range is also available (e.g., *robust_scale* function in *Scikit-Learn*). Data scaling is optional due to tree-based ensemble models being highly robust to skewed data distributions and outliers, and use of the original data distribution makes the interpretation of the models straightforward.

Another factor that can cause technical constraints in ML pipelines by increasing computational costs and through the presence of noise and multicollinearity, is high dimensionality, which is common in some ML datasets (e.g., multi-omics). This is addressed in MAIT by applying feature selection using minimum redundancy maximum relevance (mRMR) algorithm[14] to select the most relevant features based on the development set.

## Binary classification

Binary classification simplifies decision making by providing two possible prediction outcomes (e.g., survival vs. death). MAIT includes a selection of seven algorithms for binary classification: (1) LightGBM, (2) CatBoost, (3) Random Forest[15], (4) HGBC, (5) QLattice[16], as well as (6) Logistic Regression, and (7) (Gaussian) Naïve Bayes. The list of algorithms predominantly consists of tree-based ensemble models (1-4), renowned for their versatility and high performance. Simpler models, i.e., Naïve Bayes and Logistic Regression, are used as statistical alternatives to the tree-based ensemble models for performance comparison. QLattice, a more recent algorithm, can provide higher transparency than the other algorithms by producing a closed-form mathematical equation and a block diagram for visualizing feature interactions related to the outcome variable.

The models include free parameters, i.e., hyperparameters, which require configuration and tuning for optimal performance[17]. Hyperparameters, (e.g., tree depth in ensemble models, L1 regularization in logistic regression[18], etc.) are important for reasons such as regularization to counteract overfitting[18]. Certain parameters are particularly sensitive to variations in dataset characteristics, including the number of features and samples, and class distribution. Typically, these parameters are determined by exploring a parameter space using hyperparameter tuning methods or are set manually based on dataset attributes. MAIT allows hyperparameter tuning via random search and cross validation where the search space is adaptive to dataset characteristics. Hyperparameter tuning can be iterated multiple times. The cross-validation procedure is nested within the overarching cross-validation framework used for comparing different models. Evaluation metrics such as area under the receiver operating characteristic curve (AUC) and precision-recall curve (PR-AUC) are employed during hyperparameter tuning to assess model performance.

For model performance assessment and ensuring consistency across subsets of the development set, cross-validation is conducted, reporting the mean and standard deviation of various evaluation metrics. The model exhibiting the highest average performance across folds, as determined by multiple classification metrics is selected for further analysis. Relying solely on a single performance measure for model selection can introduce bias, particularly in the context of imbalanced datasets[19]. Therefore, the cross-validation procedure is stratified by the outcome variable to address class distribution imbalances and minimize bias in classification tasks.

MAIT employs various techniques to handle class imbalance [20] where one class is substantially more prevalent than the other. Sample weights are computed based on class proportions and utilized in the models to ensure balanced learning and prevent biased learning outcomes. Model performance for binary classification is evaluated using multiple metrics[21] to mitigate the risk of biased assessment associated with a single metric, and includes computation of several metrics: AUC, PR-AUC, MCC, positive predictive value, negative predictive value, sensitivity, specificity, F1 score, balanced accuracy, and Brier score. These metrics are recognized as robust measures for evaluating binary classification models[22]. The best performing model, determined by the highest grand average of MCC, AUC, and PR-AUC, is subsequently trained on the entire development set and evaluated on an unseen test set to assess its generalizability. Visualizations of ROC curves, precision recall curve, and confusion matrices enrich the evaluation process. Moreover, oversampling methods are available in MAIT to further address class imbalance.

A core component of MAIT to improve its utility for clinical research is the utilization of methods to assess model behavior, prominently featuring SHAP and other compatible methods tailored to each model type. Four methods for model interpretation are available in MAIT: SHAP, permutation-based feature importance, tree-based feature importance, and QLattice-specific feature diagram. Feature importance is estimated for each model during cross-validation and on the test set if data splitting is employed. Feature importance based only on correctly classified samples is also available in MAIT and cluster analysis based on SHAP values of features can be employed to uncover unique patient profiles [23]. As a functionality of MAIT, feature importance is depicted for each of the detected clusters as well as the model performance for the clusters. A unique capability of MAIT is the enhancement of SHAP plots by displaying feature categories.

MAIT provides clinical utility assessment through cost-sensitive decision curve analysis for binary classification tasks by comparing the net benefit[24] of a selected model to alternative models, extreme cases (e.g., all negatives), and random decisions. Additionally, MAIT enables model calibration using isotonic regression and conformal predictions, which is recommended when there is sufficient data as the approach requires an extra data split for a calibration set. MAIT also provides the option to optimize the probability threshold (cut-off value between 0 and 1) for classification models. In this scenario, the mean value of predicted probabilities for both classes is calculated for each fold in cross-validation, with the average used as an empirical estimation to approach the optimal probability threshold. The probability threshold used in each fold's model performance calculation is the median of estimated probability thresholds from previous folds, with the initial threshold value set as the proportion of samples from the minority class in the total number of samples. An additional option (model uncertainty reduction module) is proposed in MAIT to discard uncertain predictions determined by SHAP values and predicted probabilities in close proximity to the chance (neutral) level.

## Survival analysis

Many datasets contain information suitable for survival analysis, particularly evident in longitudinal data from cohort studies. Such datasets often feature a baseline date (e.g., transplant time) and an event occurring subsequent to this date (e.g., post-transplant complications or death)[5]. In these scenarios, the dataset may include a time-to-event variable spanning a follow-up period, facilitating the development and evaluation of survival models implemented by the *scikit-survival* package[25], such as random survival forest (RSF) model and Cox's proportional hazards model (CPH) with elastic net penalty.

A typical task formulation involves aggregating observations from the same patients, such as for the first month post-transplant, to predict the likelihood of developing complications within a subsequent observation period, e.g., one-year post-aggregation. Patients experiencing events beyond the observation period are censored, while those within the period have their time-to-event recorded from baseline to event date. Patients without events are assigned the time-to-event as the time to their last follow-up within the observation period.

Performance comparison of the RSF and CPH survival models typically relies on concordance index (CI) and integrated Brier score (IBS)[26]. Translation of survival model results to binary classification, as proposed in MAIT, involves generating cumulative hazard curves over time for all samples in the development set. The median cumulative hazard for each class (e.g., event and event-free) is computed, and the Euclidean distance of test set samples' cumulative hazard curves to these medians determines class assignment. This approach enables the interpretation of survival model outputs as predicted cumulative hazard curves transformed into predicted binary classes where the prediction performance of the survival model can be presented as a confusion matrix like a binary classification model.

Additionally, time-updated AUC and predicted risk scores are reported, alongside visualization of cumulative hazard curves. The survival modeling component is considered supplementary to binary classification when time-to-event data is available, enhancing the pipeline's analytical scope. RSF is favored over CPH for its ability to capture nonlinear information and feature interactions, and is interpreted by default, using feature permutation and SHAP methods via *SurvSHAP(t)* [27]. A great advantage of the survival models compared to the binary classification models is that censoring is addressed without making assumptions.

## Regression analysis

Regression analysis is a common approach used for predicting continuous outcome variables. In this context, two models are available: the Random Forest Regressor and Linear Regression. Both models are widely recognized and utilized. However, the Random Forest Regressor is typically expected to outperform the Linear Regression model in most scenarios, thus it is prioritized for explanation using the SHAP method.

## Benchmarking report and exporting

Upon completion of a pipeline run, the user will be provided with a detailed report in an HTML file and a folder (directory) containing all figures and tables generated in the user-specified analyses. The HTML file includes the outputs of all blocks of executed codes including figures describing models and datasets as well as tables featuring the values in benchmarking models (e.g., cross validation) that could easily be reviewed and compiled for publications.

## Software framework

MAIT is formatted as a monolithic notebook with sequential operations. The current version of MAIT offers a stable implementation compatible with both Linux and Windows operating systems. Additionally, it is distributed as a Docker image, enhancing its stability and reproducibility. The development strategy of MAIT prioritized simplicity for ease of comprehension and customization. Users can execute MAIT independently or within another Jupyter Notebook, leveraging tools like the *papermill* package for configuring multiple parameters. Customizability extends to the selective inclusion or exclusion of features to optimize runtime, such as limiting the number of models or bypassing hyperparameter tuning. To maximize efficiency, MAIT employs vectorization and parallelization techniques, with GPU utilization enabled for compatible models like CatBoost. Parallel processing is implemented for several functions to speed up bootstrapping methods for correlation analyses (point-biserial and mutual information), hyperparameter tuning in QLattice model, and extracting SHAP values for survival models. The suite of models and methods within MAIT is designed to accommodate a broad range of ML studies focusing on tabular data. Successful implementation of ML analyses relies on robust, well-documented Python packages for ML, statistical analyses, visualization, and model interpretation, ensuring a solid foundation for future updates and cross-platform compatibility. The most important advantage of MAIT's monolithic notebook form is that it provides a self-contained, reproducible workflow, making it easy to follow the entire process from start to finish at the cost of complex scalability as it grows.

## Tutorials and practical examples.

We have curated four tutorials based on four medical datasets that provide practical use cases for MAIT. The tutorials include examples of how to conduct studies based on classification, regression, and survival analyses. One of the tutorials is based on the Wisconsin Breast Cancer dataset[28] that includes features computed from digitized images of fine needle aspirate of breast masses, with the primary purpose of the ML task being classification of breast tumors as benign or malignant. The tutorial on the Dementia Prediction Dataset[29] presents an example of a binary classification task when both categorical and continuous features are available. In addition, we provided two tutorials based on two datasets that include genetic data on antibiotic resistance[30]. The tutorials demonstrate the strengths and versatility of MAIT and provide examples of robust prediction models. Each tutorial's dataset details and results are accessible on our GitHub repository, allowing users to select the most relevant example to guide their machine learning study needs (see Figure *2* as examples of MAIT outputs for different analyses).

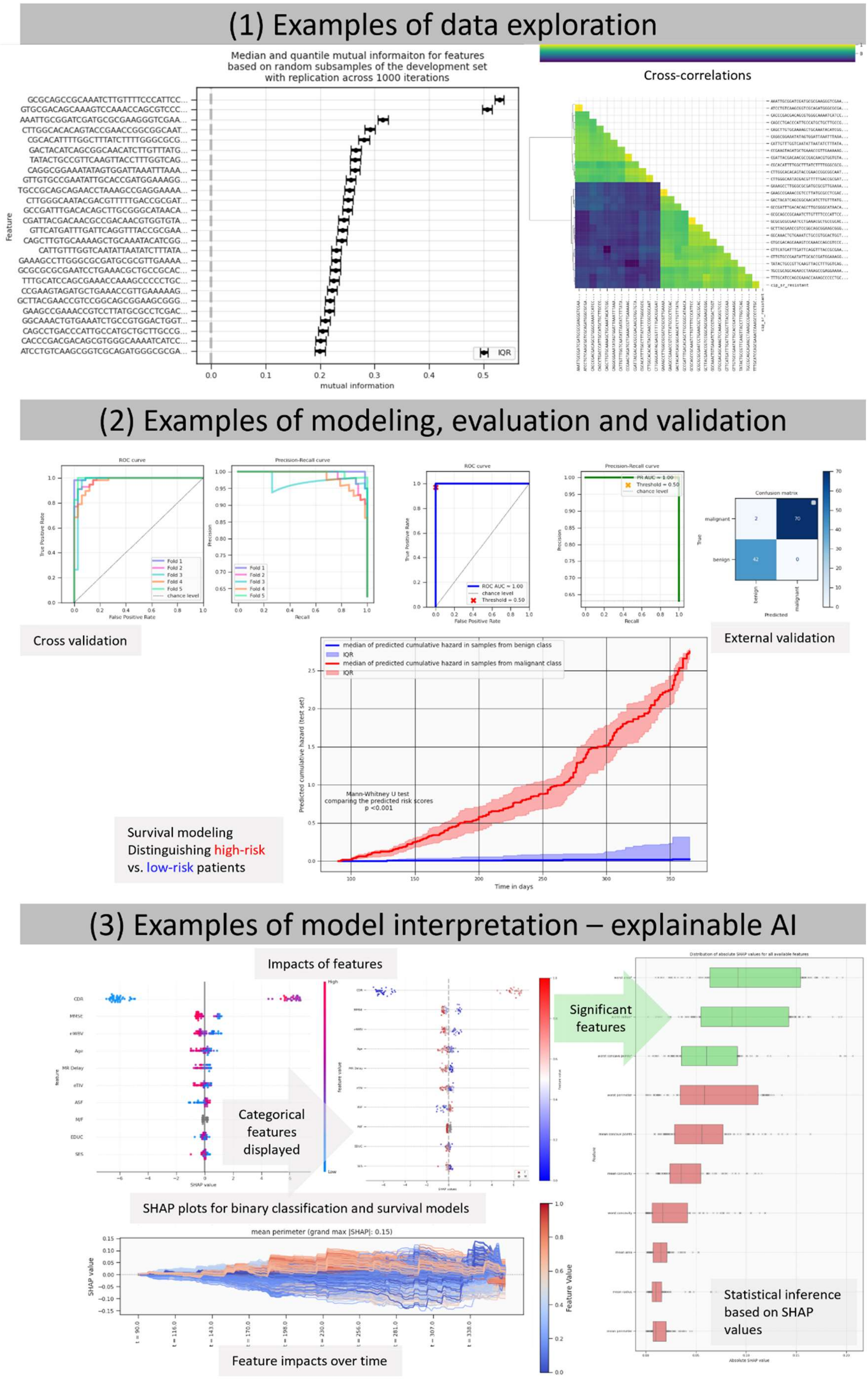

(1) Examples of data exploration
Median and quantile mutual informaiton for features based on random subsamples of the development set with replication across 1000 iterations
Feature
mutual information
IQR
Cross-correlations
(2) Examples of modeling, evaluation and validation
ROC curve
Precision-Recall curve
Cross validation
Confusion matrix
External validation
median of predicted cumulative hazard in samples from benign class
IQR
median of predicted cumulative hazard in samples from malignant class
IQR
Mann-Whitney U test comparing the predicted risk scores p <0.001
Predicted cumulative hazard (test set)
Time in days
Survival modeling
Distinguishing high-risk vs. low-risk patients
(3) Examples of model interpretation – explainable AI
Impacts of features
Significant features
Categorical features displayed
SHAP plots for binary classification and survival models
mean perimeter (grand max |SHAP|: 0.15)
SHAP value
Feature Value
Feature impacts over time
Statistical inference based on SHAP values

Figure 2 Examples of outputs from tutorials using MAIT. A subset of output figures is depicted from each analytical step as a result of conducting machine learning analyses using MAIT. The process begins with data exploration where figures display the strength of associations between variables, and is followed by model validation and evaluation exemplified by cross validation results. Finally, the associations and impacts of the variables (features) for each model type are illustrated by SHAP plots.

## Strengths and limitations of MAIT

An overview of MAIT's features is outlined in Table 1, and briefly describes key strengths and limitations of MAIT. This overview can guide researchers on the suitability of MAIT for their study purposes in comparison to AutoML frameworks. The MAIT vs AutoML comparison section lists several novel approaches incorporated into MAIT to resolve bottlenecks in ML analysis with focus on explainable AI.

Table 1 overview of scientific and technical features of MAIT. Unique functionalities provided in MAIT are outlined in the list, as well as its limitations and comparison with AutoML frameworks in general terms.

| **Features** | **Description** |
|---|---|
| Strengths of MAIT | **Data Quality & Robustness**<br>• Handles imbalanced data and incomplete datasets<br>**Explainability & Transparency**<br>• Model-specific and model-agnostic explainability<br>**Standardization & Compliance**<br>• Standardizes ML pipelines – compatible with TRIPOD+AI reporting framework<br>**Comprehensive Evaluation**<br>• Multi-metrics performance measurements and optimization<br>**Adaptability & Efficiency**<br>• Adaptive to data characteristics – e.g., for hyperparameter tuning and visualization<br>**Reliability Enhancement**<br>• Model uncertainty reduction module – to discard uncertain predictions<br>**Effective Communication**<br>• Transparent and convenient reporting of scientific and technical ML results<br>**Research & Development Flexibility**<br>• Customizable – several selective options for experimentation and hypotheses testing<br>**Broad Applicability**<br>• Simultaneous support for binary classification, regression, and survival models<br>**Feature Analysis**<br>• Unified feature importance (normalization of feature scores from multiple methods)<br>• Explores interactions between pairs of features based on SHAP and permutation methods<br>**Evaluation Scope**<br>• Ubiquitous analyses on multiple datasets (training, testing, and external validation)<br>**Healthcare & Time-to-Event Applications**<br>• Translation of survival curves to binary classifications<br>• Cost-sensitive model evaluation for clinical utility<br>**Advanced Analytics Capabilities**<br>• Statistical inference in significance of feature importance<br>• Model interpretation on all levels (population, subgroups, and individuals)<br>• Multi-level model interpretation (e.g., within cross-validation, after excluding incorrect predictions, within data-driven subgroups) |
| Limitations of MAIT | • Does not handle multi-class classification directly<br>• Does not support multi-task learning and causal ML<br>• Does not include deep learning |
| MAIT vs. AutoML | • More efficient than AutoML usually rely on testing all available configuration<br>• More interpretable than AutoML as they usually provide very limited options<br>• More customizable than AutoML for manual modifications |
| | • Less prone to bias than AutoML that tend to produce misleading results without expert supervision |

An important focus of MAIT is the derivation of feature importance. Multi-level model interpretation is used to determine feature importance for different subsets of the data (e.g., data-driven clusters based on SHAP values[23]) and for different conditions (e.g., using only correct predictions in binary classification). Multi-level model interpretation contributes to guided determination of feature importance as opposed to, for instance, using the whole dataset to determine feature importance without consideration that samples with incorrect predictions could be adding irrelevant information to complicate interpretation. In addition, MAIT provides a significance test using bootstrap subsampling (95% confidence) that identifies statistically significant features based on their SHAP values, marking those that have an Interquartile Range (IQR) crossing a data-driven threshold less than 5% of the time. Furthermore, the availability of the other methods to determine feature importance like the permutation method[15], strengthens model interpretation on different levels (e.g., within cross validation and on the test set). MAIT also analyzes feature interactions using two methods: SHAP values, and a feature permutation method, which calculates importance scores for all possible feature pairs. These are key contributions of MAIT in addition to rich visualizations that can reveal important aspects of models and the underlying data, addressing shortcomings of existing software solutions (e.g., SHAP plots with the presence of categorical features).

To the best of our knowledge, MAIT provides a unique solution to unify and compare findings from a survival model (time-to-event outcome variable) and its binary classification counterpart (when the outcome variable is defined as binary classes) for the same dataset. It translates predicted cumulative hazard curves to binary classifications by comparing the difference of the predicted curves from a test set with the median curves representing each of the two classes (based on non-censored samples from the training set). The translation to binary classification is useful for decision-making processes where a predicted cumulative hazard curve alone is less interpretable and actionable. This also allows for comparing feature importance between the two models of different types.

AutoML frameworks (like H2O AutoML[31]), an alternative to MAIT, boast ease of use but suffer from significant limitations[32]. They incur high computational costs due to exhaustive method and model testing, often yielding negligible performance differences. In contrast, MAIT optimizes resource utilization by allowing users to specify desired resources and providing a curated set of methods and models with user guidance. Beyond computational efficiency, AutoML frameworks are criticized for their limited customizability, lack of model transparency, and overemphasis on top-performing models without regard for interpretability—a crucial aspect in clinical research. MAIT addresses these concerns by supplying an open-source, modifiable Jupyter Notebook with configurable pipeline options, and integrated model interpretation methods for clearer reasoning explanations. In AutoML frameworks, there is a risk of overlooking important details such as imbalanced class distribution and stratified data splitting. In contrast, MAIT employs a multi-measure approach to robust validation, offers data imputation and encoding for incomplete datasets, and contributes to model generalizability through adaptive hyperparameter tuning.

An alternative approach to AutoML is to design and implement a new pipeline from scratch, which could be inefficient and prone to errors. Instead, MAIT presents a standardized, extensible framework based on ML best practices [1,21,33,34]. Notably, MAIT uniquely supports simultaneous development of binary classification, regression, and survival models derived from a single dataset whilst supporting external validation. This feature allows direct head-to-head comparisons across models that are otherwise almost infeasible (due to technical requirements including identical computerized randomizations and software environments). This capability is facilitated by using the innovative translations of predicted cumulative hazard curves to binary classifications for comparable predictions across the two model types.

MAIT advocates for the human-in-the-loop formalism[7] while maintaining a balance between arbitrary ML designs and autopilot approaches based on AutoML frameworks for high quality research. The efficient set of models and methods with selective options and room for customisation are the key to approach this aim. The MAIT framework introduces novel synergies, such as combining feature importance rankings (SHAP, permutation, tree-based) and a distance-based approach for survival to binary predictions. Method selections prioritize efficiency and stability, such as mRMR for feature selection and kNN for data imputation, with thoughtful omissions (e.g., SMOTE[35], MICE[36]) to maintain simplicity and performance.

## Conclusion

MAIT provides a novel framework for explainable machine learning (ML) research on tabular datasets, synergistically integrating established methodologies with cutting-edge algorithms and novel approaches for improved visualizations to interpret models, supporting statistical inference for improved confidence in feature importance derivations, and model validation. By presenting a standardized, adaptable, and user-friendly platform, MAIT accelerates the exploration of complex datasets, facilitates the development of robust and deployable prediction models tailored to diverse patient profiles. MAIT is a powerful tool with potential applications in enhancing the translational relevance of ML insights in clinical settings and driving advancements in personalized medicine and evidence-based healthcare decision-making.

## Funding

This work was supported by the Danish National Research Foundation [DNRF126].

## Data availability

Examples of execution of MAIT on publicly available datasets are available at
https://github.com/PERSIMUNE/MAIT.

## Conflict of interest

None declared.

## Code availability

The source codes for MAIT are publicly available at https://github.com/PERSIMUNE/MAIT.

## Acknowledgements

We would like to thank Henrik Lassen at Centre of Excellence for Health, Immunity and Infections (CHIP), Rigshospitalet, Copenhagen University Hospital, Denmark for his help in maintaining CHIP HPC servers, that were the computing infrastructure used to develop and test the pipeline.